\documentclass[journal]{IEEEtran}
\usepackage{bm}
\usepackage{amsmath}
\usepackage{amssymb}
\usepackage{amsthm}
\usepackage{mathrsfs}
\usepackage{enumerate}
\usepackage{multirow}
\usepackage{color}
\usepackage{subfigure}
\usepackage[square, comma, sort&compress, numbers]{natbib}
\usepackage[pagebackref=false,breaklinks=true,letterpaper=true,colorlinks,bookmarks=false]{hyperref}
\usepackage{times}
\usepackage{breakurl}
\usepackage{array}
\usepackage{verbatim}
\usepackage{algorithm}
\usepackage{algpseudocode}

\makeatletter
\def\hlinewd#1{%
  \noalign{\ifnum0=`}\fi\hrule \@height #1 \futurelet
   \reserved@a\@xhline}
\makeatother

\ifCLASSINFOpdf
  \usepackage[pdftex]{graphicx}
  \usepackage{graphics}
  \usepackage{graphicx}
  \usepackage{epsfig}
  \usepackage{epstopdf}
  \graphicspath{{./fig/}}
  \DeclareGraphicsExtensions{.pdf,.jpeg,.png,.eps}
\else
  \usepackage[dvips]{graphicx}
  \graphicspath{{./fig/}}
  \DeclareGraphicsExtensions{.eps}
\fi
\begin{document}

%
\title{Deep Hybrid Similarity Learning for Person Re-identification}

%
%
%


\author{Jianqing Zhu, Huanqiang Zeng, Shengcai Liao, Zhen Lei, Canhui Cai, and LiXin Zheng

\thanks{This work was supported in part by the Scientific Research Funds of Huaqiao University under the Grant 16BS108, in part by the National Natural Science Foundation of China under the Grants 61602191, 61375037, 61473291, 61572501, 61502491, 61572536, 61372107 and 61401167, in part by the Natural Science Foundation of Fujian Province under the Grant 2016J01308, in part by the Opening Project of Sate Key Laboratory of Digital Publishing Technology under the Grant FZDP2015-B-001, in part by the Zhejiang Open Foundation of the Most Important Subjects, in part by the High-Level Talent Project Foundation of Huaqiao University under the Grants 14BS201 and 14BS204.
}
\thanks{Jianqing Zhu, Canhui Cai and LiXin Zheng are with the College of Engineering, Huaqiao University, Quanzhou, Fujian 362021, China and Fujian Provincial Academic Engineering Research Centre in Industrial Intellectual Techniques and Systems (e-mail: \{jqzhu, chcai\}@hqu.edu.cn and zlxgxy@qq.com).}
\thanks{Huanqiang Zeng (corresponding author) is with the College of Information Science and Engineering, Huaqiao University, Xiamen, Fujian 361021, China (e-mail: zeng0043@hqu.edu.cn).}
\thanks{Shengcai Liao and Zhen Lei are with the Center for Biometrics and Security Research and National Laboratory of Pattern Recognition, Institute of Automation, Chinese Academy of Sciences, Beijing 100190, China (e-mail: \{scliao, zlei\}@cbsr.ia.ac.cn).}
}

%
%

\markboth{}%
{Shell \MakeLowercase{\textit{et al.}}: Bare Demo of IEEEtran.cls
for Journals}
%



\maketitle

\begin{abstract}

  Person Re-IDentification (Re-ID) aims to match person images captured from two non-overlapping cameras. In this paper, a deep hybrid similarity learning (DHSL) method for person Re-ID based on a convolution neural network (CNN) is proposed. In our approach, a CNN learning feature pair for the input image pair is simultaneously extracted. Then, both the element-wise absolute difference and multiplication of the CNN learning feature pair are calculated. Finally, a hybrid similarity function is designed to measure the similarity between the feature pair, which is realized by learning a group of weight coefficients to project the element-wise absolute difference and multiplication into a similarity score. Consequently, the proposed DHSL method is able to reasonably assign parameters of feature learning and metric learning in a CNN so that the performance of person Re-ID is improved. Experiments on three challenging person Re-ID databases, QMUL GRID, VIPeR and CUHK03, illustrate that the proposed DHSL method is superior to multiple state-of-the-art person Re-ID methods.
\end{abstract}



%
\IEEEpeerreviewmaketitle

\section{Introduction}\label{sec:intro}
\begin{figure}[tp]
    \centering
    \includegraphics[width=0.95\linewidth]{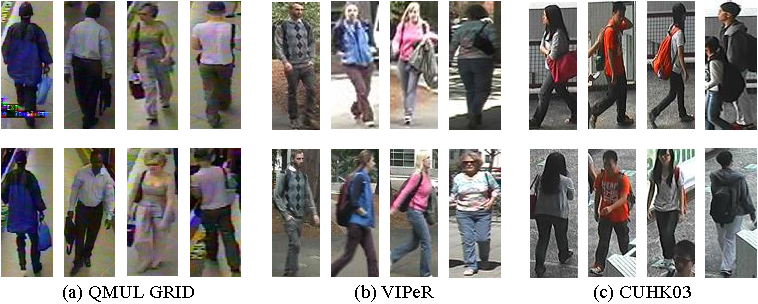}
    \caption{Example pairs of images from the QMUL GRID \cite{grid}, VIPeR \cite{viper} and CUHK03 \cite{cuhk03} databases. Images in the same column represent the same person captured under different cameras.}
    \label{fig:example}
\end{figure}

 Person Re-IDentification (Re-ID) plays an important role in video surveillance for public safety \cite{grid}. However, it is a challenging problem since person images are usually with low resolution and partial occlusion and contain large intra-class variations of illumination, viewpoint and pose. See Fig. \ref{fig:example} for some person image examples. Terefore, how to develop an effective person Re-ID method becomes a very desirable topic.


 The two fundamental problems critical for person Re-ID are feature representation and similarity metric. For feature representation, there are Fisher vectors (LDFV) \cite{fishervec}, bio-inspired features (kBiCov) \cite{kBiCov}, symmetry-driven accumulated local features (SDALF) \cite{sdalf}, structural constraints enhanced feature accumulation (SCEFA) \cite{scefa}, color invariant signature \cite{color}, salience matching \cite{salienceiccv,saliencecvpr}, ensemble of local features (ELF16) \cite{elf16}, mid-level learning features \cite{midlevelfeature} and convolutional neural network learning features \cite{crosscnn}, and so on.

 For similarity metrics, many machine learning algorithms are developed to calculate the similarity between a person image pair, such as ranking support vector machine (Ranking SVM) \cite{ranksvm}, partial least square (PLS) \cite{pls}, Boosting \cite{boosting}, multi-task learning \cite{mtml,mtllr}, metric learning \cite{itml,lmnn,kissme,rdc,rpm,ladf,lomo,psd,me} and convolutional neural networks (CNNs) \cite{dml,improvedcnn,deepfeature,personnet,edm,can,deeprank,gscnn,ssdal,dgd}. Note that the metric learning and CNN based person Re-ID methods are the most popular methods, which will be highlighted as follows.

 The metric learning based person Re-ID methods \cite{rdc,rpm,ladf,lomo,psd,me} learn a matrix with $d \times d$ parameters to calculate the Mahalanobis distance between two $d$ dimensional hand-crafted features as the similarity measurement between two person images. However, they are prone to over-fitting on a small database. Because the number of parameters in a learned Mahalanobis matrix is square of the feature dimension and this is a large number. For this, the principal component analysis (PCA) is usually used for feature dimension compression before metric learning. However, it is not optimal for metric learning, since PCA is not jointly optimized with metric learning. Another solution for feature dimension compression is proposed in \cite{lomo}, which is able to jointly learn a discriminant low dimensional subspace and a similarity metric by the cross-view quadratic discriminant analysis. In \cite{lomo}, although the number of parameters in the projection matrix is reduced to be $d \times s$, it is still large, where $s$ is compressed feature dimension.

 Most existing CNN based person Re-ID methods \cite{dml,deepfeature,deeprank,dgd,edm,can,gscnn,ssdal} undervalue the similarity learning, which only apply the simple cosine or Euclidean distance function to measure the similarity between a CNN learning feature pair.
 Moreover, recently CNN based person Re-ID methods \cite{dgd,can,gscnn} pay more attentions on the feature learning with a deeper feature learning module. In addition, several specified layers are designed for further emphasizing feature learning, such as attention component \cite{can} and matching gate architecture \cite{gscnn}. The tendency of undervaluing similarity learning and emphasizing feature learning leads to two deficiencies. (1) The simple cosine or Euclidean distance function is not very discriminative for measuring the similarity between a CNN learning feature pair. (2) The deeper feature learning module is, the larger scale training dataset will be required. As a result, there is still ample room for the improvement on CNN based person Re-ID methods.

 In this paper, an effective deep hybrid similarity learning (DHSL) method for person Re-ID is proposed. As a CNN based person Re-ID method, the major contribution of this paper is to improve person Re-ID performance by reasonably assigning complexities of metric learning and feature learning modules in the CNN model. For the metric learning module, a hybrid similarity function with reasonable parameters is designed to measure person similarity. For the feature learning module, a light convolution neural network only with three convolution layers is applied to extract features. The hybrid similarity function is realized by learning a group of weight coefficients to project the element-wise absolute difference and multiplication of a CNN learning feature pair into a similarity score. Since both the element-wise difference and multiplication of a CNN learning feature pair are considered, the hybrid similarity function is more discriminative than the simple cosine or Euclidean distance based similarity metric. Note that the number of parameters in the hybrid similarity function is only 2 times of the feature dimension, which is much small than that of the Mahalanobis distance based similarity metric. Consequently, it has been verified from extensive experiments on three challenging person Re-ID databases (i.e., QMUL GRID \cite{grid}, VIPeR \cite{viper} and CUHK03 \cite{cuhk03}) that the proposed DHSL method outperforms state-of-the-art person Re-ID methods.

 The rest of this paper is organized as follows. Section \ref{sec:method} introduces the details of the proposed deep hybrid similarity learning method. Section \ref{sec:exp} presents experimental results and analyses. Section \ref{sec:conclusion} concludes this paper.

\begin{figure}[t]
    \centering
    \includegraphics[width=1\linewidth]{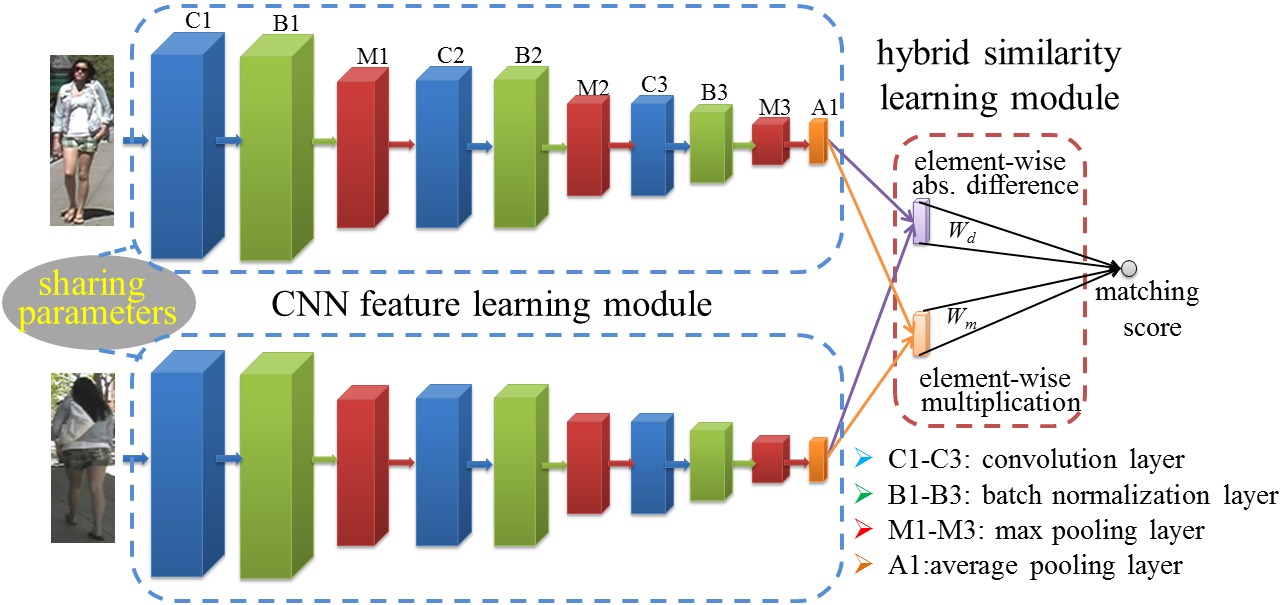}
    \caption{The framework of the proposed DHSL method for person Re-ID.}
    \label{fig:framework}
\end{figure}

\section{Deep Hybrid Similarity Learning (DHSL) for Person Re-ID}\label{sec:method}
 Fig. \ref{fig:framework} shows the framework of the proposed DHSL method. Since the main novelty is embodied in the hybrid similarity learning module, the hybrid similarity learning module will be first introduced and then the details of the CNN feature learning and the objection function construction will be presented.

\subsection{Hybrid Similarity Learning Module}\label{sec:csml}
 Assuming that the feature pair produced by the CNN feature learning module is ${X_1,X_2}\in{\Re ^d}$. Now, the question boils down to the similarity measurement between a feature pair. In this work, by analyzing the relationship among the Mahalanobis, Euclidean and cosine distances, a hybrid similarity function learned on the element-wise absolute differences and multiplications of feature pairs is proposed as below.

 Firstly, the Mahalannobis and Euclidean distances are formulated in Eq. (\ref{equ:mdist}) and Eq. (\ref{equ:eudist}), respectively.
\begin{equation}\label{equ:mdist}
\small
\begin{array}{l}
{{d}}_{{M}}^2({X_1},{X_2}) = {({X_1} - {X_2})^T}M({X_1} - {X_2})\\
~~~~~~~~~~~~~~~=(vec(M))^Tvec((X_1-X_2)(X_1-X_2)^T).
\end{array}
\end{equation}
\begin{equation}\label{equ:eudist}
{{d}}_{{E}}^2({X_1},{X_2}) = {({X_1} - {X_2})^T}({X_1} - {X_2}).
\end{equation}

 The $vec(\cdot)$ function in Eq. (\ref{equ:mdist}) is used to rearrange a $d\times d$ dimensional matrix into a $d^2 \times 1$ dimensional vector. From these two equations, one can see that the Euclidean distance is the summation of $d$ square differences, in which the $k$-th square difference is calculated at the $k$-th dimension. On the contrary, the Mahalanobis distance includes not only a linear combination of the $d$ square differences, but also a linear combination of $d^2-d$ correlations of the differences and each correlation is calculated between a feature pair at different dimensions. Hence, it can be easily observed that the Mahalanobis distance formulation is much more complex than the Euclidean distance formulation.

 In addition to the Mahalannobis and Euclidean distances, the cosine distance is also one of commonly-used similarity metrics. Assuming that $X_1$ and $X_2$ have been $\ell_2$ normalized, then cosine distance between them can be formulated as follows:
\begin{equation}\label{equ:cosdist}
  d_{cos}({X_1},{X_2}) = X_1^T\cdot X_2=(X_1.*X_2)^T\cdot{\rm \textbf{1}},
\end{equation}
 where ${\rm \textbf{1}}=[1,1,...,1]^T \in{\Re ^d}$ is a constant vector. To take a deep insight to the cosine distance, the Mahalanobis distance in Eq. (\ref{equ:mdist}) is further expanded as follows:
\begin{equation}\label{equ:mdist2}
\small
\begin{array}{l}
{{d}}_{{M}}^2({X_1},{X_2})=(vec(M))^Tvec((X_1-X_2)(X_1-X_2)^T)\\
=(vec(M))^Tvec(X_1X_1^T+X_2X_2^T-X_1X_2^T-X_2X_1^T).
\end{array}
\end{equation}

 From Eqs. (\ref{equ:cosdist}) and (\ref{equ:mdist2}), one can see that the cosine distance is the summation of $d$ correlations, in which the $k$-th correlation is only for the cross-correlation between $X_1$ and $X_2$ at the $k$-th dimension. On the contrary, the Mahalanobis distance considers the cross-correlation of $X_1$ and $X_2$ and that of $X_2$ and $X_1$, the self-correlation of $X_1$ and that of $X_2$, in which both cross-correlations and self-correlations are calculated at the same dimension and different dimensions. Therefore, it can be concluded that the cosine distance is a simplification of the Mahalanobis distance.
\begin{table}[t]
\centering
\caption{The numbers of parameters in the Mahalannobis distance, the Euclidean distance, the cosine distance and the proposed hybrid similarity, where $d$ is the feature dimension.}\label{tab:computation}
\setlength{\tabcolsep}{1.8pt}
\begin{tabular}{c|c|c|c|c}
\hlinewd{1.5pt}
   Methods & \begin{tabular}{c}
             Mahalannobis \\
             Eq. (\ref{equ:mdist}) \\
            \end{tabular}&
            \begin{tabular}{c}
             Euclidean \\
             Eq. (\ref{equ:eudist}) \\
            \end{tabular}&
            \begin{tabular}{c}
             Cosine \\
             Eq. (\ref{equ:cosdist}) \\
            \end{tabular}&
            \begin{tabular}{c}
             Hybrid similarity \\
             Eq. (\ref{equ:hybrid}) \\
            \end{tabular}\\
\hlinewd{0.8pt}
   Parameter number  & $d\times d$ & 0 & 0 & $2d$ \\
\hlinewd{1.5pt}
\end{tabular}
\vspace{-.4cm}
\end{table}

Based on the above analysis, the Euclidean or cosine distance functions that are commonly-used in CNN are simpler but with lower discriminative ability to measure the similarity between a CNN learning feature pair. While the Mahalanobis distance is more discriminative but requires a large number of parameters ($d\times d$), which is thus not suitable to be integrated with a CNN directly. Therefore, we propose a hybrid similarity function that has not only a strong discriminative ability but also a reasonable number of parameters. The proposed hybrid similarity function is a linear combination of the element-wise absolute difference and multiplication between a feature pair as follows:
\begin{equation}\label{equ:hybrid}
  d_H({X_1},{X_2}) = W_d^T|X_1-X_2|+W_m^T(X_1.*X_2),
\end{equation}
 where $W_d\in{\Re^{d}}$ and $W_m\in{\Re^{d}}$ are two group of coefficients used to project the element-wise absolute difference and multiplication, respectively.

 Table \ref{tab:computation} summarizes the number of parameters in the Mahalannobis distance, the Euclidean distance, the cosine distance and the proposed hybrid similarity. Compared with the Mahalanobis distance in  Eq. (\ref{equ:mdist2}), the proposed hybrid similarity function is much simpler, since it only needs $2d$ parameters to take the difference and correlation information between pair features at each feature dimension into account. Compared with the Euclidean distance in Eq. (\ref{equ:eudist}), the proposed hybrid similarity function utilizes the absolute difference to replace the square difference for further simplifying the computation. In addition, the proposed hybrid similarity function considers both the element-wise difference and multiplication of a CNN learning feature pair by learning based coefficients $W_d, W_m\in{\Re^{d}}$, it therefore has a stronger discriminative ability.

 Furthermore, to learn the proposed hybrid similarity function, the element-wise difference and multiplication layers are designed and integrated with the CNN feature learning module, as shown in Fig. \ref{fig:framework}. The forward and backward propagations of these two layers, and the corresponding objective functions are designed as follows.

\subsubsection{Element-wise Absolute Difference Layer}
 The forward and backward propagations of the element-wise absolute difference layer are formulated as follows:
\begin{equation}\label{equ:absdiff}
  Diff({X_1},{X_2}) = |{X_1} - {X_2}|,
\end{equation}
\begin{equation}\label{equ:diff1}
\frac{{\partial Diff}}{{d{  X^i_1}}} = \left\{ \begin{array}{l}
1,{\rm{        }}if{\rm{ }}{X^i_1} > {X^i_2},\\
0,{\rm{       }}if{\rm{ }}{ X^i_1} = {X^i_2},\\
-1,{\rm{      }}otherwise,
\end{array} \right.
{\rm{and~~}}
\frac{{\partial Diff}}{{d{X^i_2}}} = -\frac{{\partial Diff}}{{d{X^i_1}}},
\end{equation}
where $X^i_1$ and$X^i_2$ represent $i$-th dimensions of $X_1$ and $X_2$, respectively.

\subsubsection{Element-wise Multiplication Layer}
 The forward and backward propagations of the element-wise multiplication layer are formulated as follows:
\begin{equation}\label{equ:multi}
  Mult({X_1},{X_2}) = {X_1}{.*}{X_2},
\end{equation}
\begin{equation}\label{equ:multi1}
  \frac{{\partial Mult}}{{d{X_1}}} = {X_2} {\rm{~and~}}  \frac{{\partial Mult}}{{d{X_2}}} = {X_1}.
\end{equation}

\subsection{CNN Feature Learning Module}
 As discussed before, the propose hybrid similarity function is more discriminative than the simple cosine or Euclidean distance based similarity metric. Therefore, we apply a light CNN feature learning module for balancing the complexities between metric learning and feature learning modules.

 As shown in Fig. \ref{fig:framework}, the proposed CNN feature learning module consists of two parameter sharing feature extraction branches. The way of sharing parameters means the parameters of each branch are the same, which can be referred in \cite{dml,deepfeature,gscnn,can}. In each branch, there are three convolution layers (i.e., C1, C2 and C3), three batch normalization \cite{bnorm} layers (i.e., B1, B2 and B3), three max pooling layers (i.e., M1, M2 and M3) and one average pooling layer (i.e., A1).

 For C1, C2 and C3 layers, the zero padding operation is applied and $3\times3$ tiny sized filters are applied for saving filter parameters. For M1, M2 and M3 layers, the $3\times 3$ max pooling operation is used. The strides of three convolution layers and three max pooling layers are set as 1 and 2, respectively. The A1 layer uses a $1\times6$ average pooling operation to calculate an average feature map on the horizontal direction to produce the feature representation of an input image. This strategy is inspired by our previous work \cite{lomo}, to improve the viewpoint robustness of the learned features. Moreover, the stride of the A1 layer is set as 1 to avoid compress features excessively. In this paper, input images are resized into $128\times48$, then a recommendation parameter configuration (e.g. C1, C2 and C3 hold 32, 64 and 128 channels, respectively) for the CNN feature learning module is applied and its details can be referred to Table \ref{tab:framework}.

\begin{table}[t]
\centering
\caption{The parameter details of the CNN feature learning module. }\label{tab:framework}
\setlength{\tabcolsep}{.4pt}
{
\begin{tabular}{c|c|c|c|c|c}
  \hlinewd{1.5pt}
  Name        & \begin{tabular}{c}
                 Output size\\
                 ($h \times w \times c$)\\
                \end{tabular}
              & Neuron
              & \begin{tabular}{c}
                Filter \\
               ($h\times w \times c \times g$) \\
                \end{tabular}
              & Stride
              & \begin{tabular}{c}
                 Pooling\\
                 operation\\
                 ($h\times w$)
              \end{tabular} \\
  \hlinewd{0.8pt}
    C1  & 128$\times$48$\times32$ & -    & 3$\times$3$\times$3$\times$32   & 1   &-  \\
\hline
    B1  & 128$\times$48$\times32$ & ReLU & -          &-    &-       \\
\hline
    M1  & 64$\times$24$\times32$  & -    & -          & 2   &3$\times$3    \\
\hline
    C2  & 64$\times$24$\times64$  & -    & 3$\times$3$\times$32$\times$64  & 1   &-  \\
\hline
    B2  & 64$\times$24$\times64$  & ReLU & -          &-    &-          \\
\hline
    M2  & 32$\times$12$\times64$  & -    & -          &2    &3$\times$3        \\
\hline
    C3  & 32$\times$12$\times128$ & -    & 3$\times$3$\times$64$\times$128 & 1   &-  \\
\hline
    B3  & 32$\times$12$\times128$ & ReLU & -          &-    &-  \\
\hline
    M3  & 16$\times$6$\times128$  & -    & -          & 2   &3$\times$3          \\
\hline
    A1  & 16$\times$1$\times128$  & -    & -          & 1   &1$\times$6           \\
\hline
    \begin{tabular}{c}
    element-wise\\
    abs. difference \\
    \end{tabular}           & 16$\times$1$\times$128 & - & - &- &-\\
\hline
    \begin{tabular}{c}
    element-wise\\
   multiplication \\
    \end{tabular}          & 16$\times$1$\times$128 & - & - &- &-\\
\hlinewd{1.5pt}
\multicolumn{6}{l}{Parameters $h$, $w$, $c$ and $g$ represent height, width, channel and group sizes,}\\
\multicolumn{6}{l}{respectively.}
\end{tabular}
\vspace{-.4cm}
}
\end{table}

\subsection{Objective Function Construction}
 Similar to \cite{dml,improvedcnn}, the person Re-ID problem is transformed into a classification problem: if a pair of person images holds the same ID, it will be a positive sample. Otherwise, it is a negative sample. To find a discriminative projection vector $W$ to ensure the classification accuracy, we apply a log-logistic model \cite{logistic} to construct the objective function for the proposed hybrid similarity function learning:
\begin{equation}\label{equ:obj}
\small
{W = \mathop {\arg \min }\limits_W \{\frac{1}{K}[\sum\nolimits_{k = 1}^K {\log (1 + {e^{ - {y_k}{W^T}{Z_k}}})}]+\frac{1}{2}\alpha||W||_2^2\},}
 \end{equation}
 where $W=[W_d,W_m]\in{\Re^{d+d}}$ is the hybrid project vector, $\alpha$ is a constant used to control the contribution of the regularization item, $\{(y_k,Z_k), k=[1,2,...,K]\}$ is the training dataset with $K$ samples, $y_{(.)}\in\{-1,1\}$ represents a class label, $Z_{(.)}=[Diff,Mult]\in{\Re^{d+d}}$ is an integration feature consisting of element-wise absolute difference ($Diff\in{\Re^d}$) and element-wise multiplication ($Mult\in{\Re^d}$). Note that this equation is optimized by using the stochastic gradient descent algorithm \cite{alexnet}.

\section{Experiment and Analysis}\label{sec:exp}
 To validate the performance, the proposed DHSL method is evaluated and then compared with multiple state-of-the-art person Re-ID methods on three challenging datasets, QMUL GRID \cite{grid}, VIPeR \cite{viper} and CUHK03 \cite{cuhk03}.

\subsection{Dataset and Evaluation Protocol}
 \textbf{QMUL GRID} \cite{grid} contains 250 pedestrian image pairs, and each pair contains two images of the same person captured from 8 disjoint camera views in a underground station. Besides, there are 775 background images that do not belong to the 250 persons and can be used to enlarge the gallery set. The experimental setting of 10 random trials is provided by the GRID dataset. For each trial, 125 image pairs are used for training, and the remaining 125 image pairs as well as the 775 background images are exploited for testing. The average of cumulative match characteristic (CMC) \cite{viper} curves calculated on the 10 random trials is employed as the final result.

 \textbf{VIPeR} \cite{viper} includes 632 person image pairs captured by a pair of cameras in an outdoor environment. Images in VIPeR have large variations in background, illumination, and viewpoint. Our experiments follow the widely adopted experimental protocol on VIPeR, which randomly divides 632 image pairs into 2 parts: half for training and the other half for testing, and repeats the procedure 10 times to obtain the average CMC as the final result.

 \textbf{CUHK03} \cite{cuhk03} has 13,164 images of 1,360 pedestrians. These images are captured by 6 cameras over months, in which each person is observed by two disjoint camera views and has 4.8 images on average in each view. Both manually labeled and auto-detected pedestrian images are provided in CUHK03. Our experiments follow the same protocol in \cite{cuhk03} as below. The CUHK03 is split into a training set with 1,160 persons and a test set with 100 persons. The experiments are conducted with 20 random trials and all the CMC curves are computed with the single-shot setting.

\begin{figure}[tp]
    \centering
    \includegraphics[width=.8\linewidth]{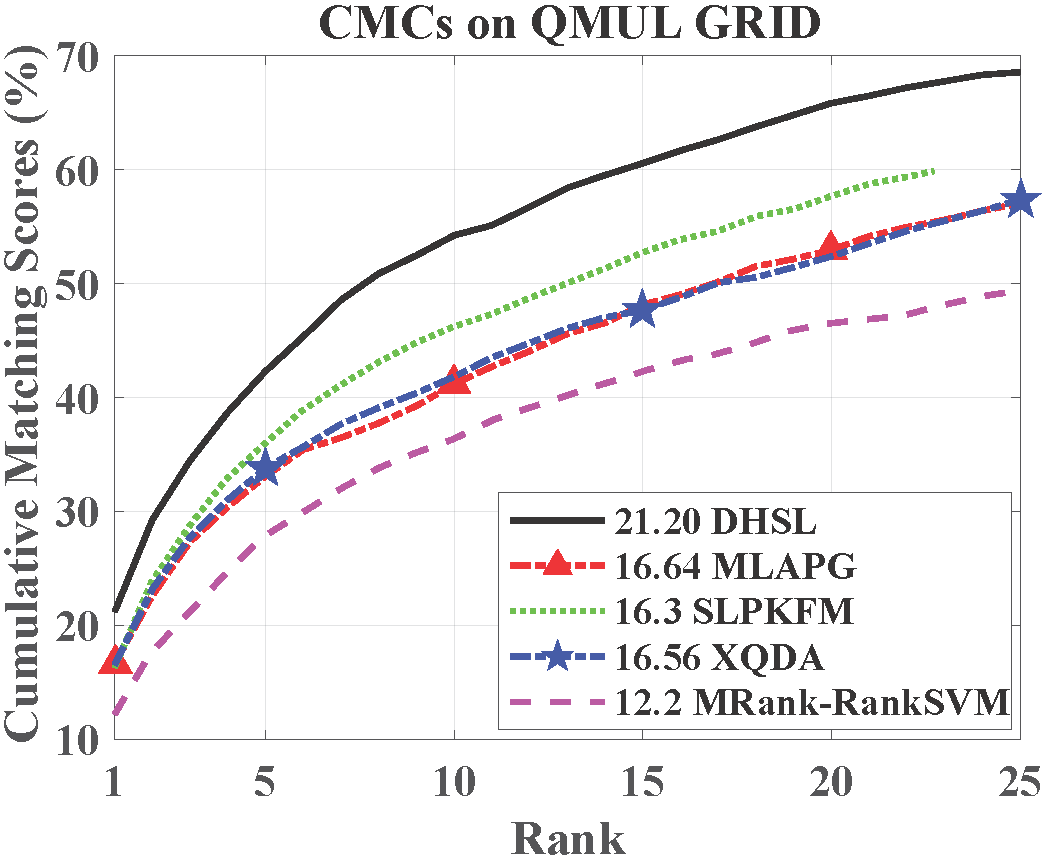}
    \caption{CMC curves and rank-1 identification rates on QMUL GRID \cite{grid} (gallery: 125 individuals+775 background images).}
    \label{fig:grid_result}
\end{figure}

\begin{table}[tp]
\centering
\caption{Performance comparison of the proposed DHSL method and multiple state-of-the-art methods on QMUL GRID \cite{grid} (gallery: 125 individuals+775 background images).} \label{tab:grid}
\setlength{\tabcolsep}{1.2pt}
\begin{tabular}{c|c|c|c|c}
\hlinewd{1.5pt}
   Method & Rank=1 (\%) & Rank=10 (\%) & Rank=20 (\%) & Reference \\
\hlinewd{0.8pt}
  DHSL             & \textbf{21.20} & \textbf{54.24}  &  \textbf{65.84}   &   proposed      \\
\hline
  SSDAL            &19.1  &45.8  &58.1  & 2016 ECCV \cite{ssdal}\\
\hline
  MLAPG            &16.64 &41.20 &52.96 & 2015 ICCV \cite{psd}\\
\hline
  SLPKFM           &16.3  &46.0  & 57.6 & 2015 CVPR \cite{kfm} \\
\hline
  XQDA             &16.56 &41.84 &52.40 & 2015 CVPR \cite{lomo}\\
\hline
  \scriptsize {MRank-RankSVM}    &12.2  &36.3  &46.6  & 2013 ICIP \cite{mrk} \\
\hlinewd{1.5pt}
\end{tabular}
\end{table}

\subsection{Implementation Detail}
 The implementation details of the proposed DHSL method can be described as below. All images in QMUL GRID, VIPeR and CUHK03 are scaled to $128 \times 48$ pixels. All the three datasets are augmented by the horizontal mirror operation. In addition, the small datasets QMUL GRID and VIPeR are further augmented by randomly rotating each image in the range $[-3^{\circ}, 0^{\circ}]$ and $[0^{\circ}, 3^{\circ}]$. For the network training, we initialize the weights in each layer from a normal distribution $N(0,0.01)$ and the biases as 0. The regularization weight $\alpha$ in Eq. (\ref{equ:obj}) for the two small datasets QMUL GRID and VIPeR is set as $5\times10^{-2}$, while that for the dataset CUHK03 is set as $5\times10^{-4}$. The size of mini-batch is 128 including 64 positive and 64 negative image pairs, and both positive and negative pairs are randomly selected from the whole training dataset. The momentums are set as 0.9. A base learning rate is started from 0.001 for QMUL GRID and VIPeR, while a larger learning rate (i.e., 0.01) is started for CUHK03 to accelerate the training process. The learning rates are gradually decreased as the training progress. That is, if the objective function is convergent at a stage, the learning rates are reduced to $1/10$ of the original values, and the minimum learning rate is $10^{-4}$. Moreover, the hard negative mining \cite{improvedcnn} is performed on CUHK03, since negative pairs on this big dataset are desirable to be used as fully as possible.

\subsection{Comparison with State-of-the-Art Methods}

\subsubsection{\textbf{Result on QMUL GRID}}
 Fig. \ref{fig:grid_result} and Table \ref{tab:grid} show the performance comparison between the proposed DHSL method and state-of-the-art person Re-ID methods on QMUL GRID \cite{grid}. It can be observed that the proposed DHSL method outperforms the semi-supervised deep attribute learning (SSDAL) \cite{ssdal} method by 2.1\% rank-1 recognition rates, without using the assistance of human attributes.
 Moreover, the proposed DHSL method consistently outperforms the state-of-the-art metric learning based methods MLAPG \cite{psd}, the polynomial kernel feature map (SLPKFM) method \cite{kfm} and XQDA \cite{lomo} at different ranks. This study shows that even for the small dataset QMUL GRID, the proposed method DHSL is able to obtain a promising performance.


\begin{figure}[tp]
    \centering
    \includegraphics[width=.85\linewidth]{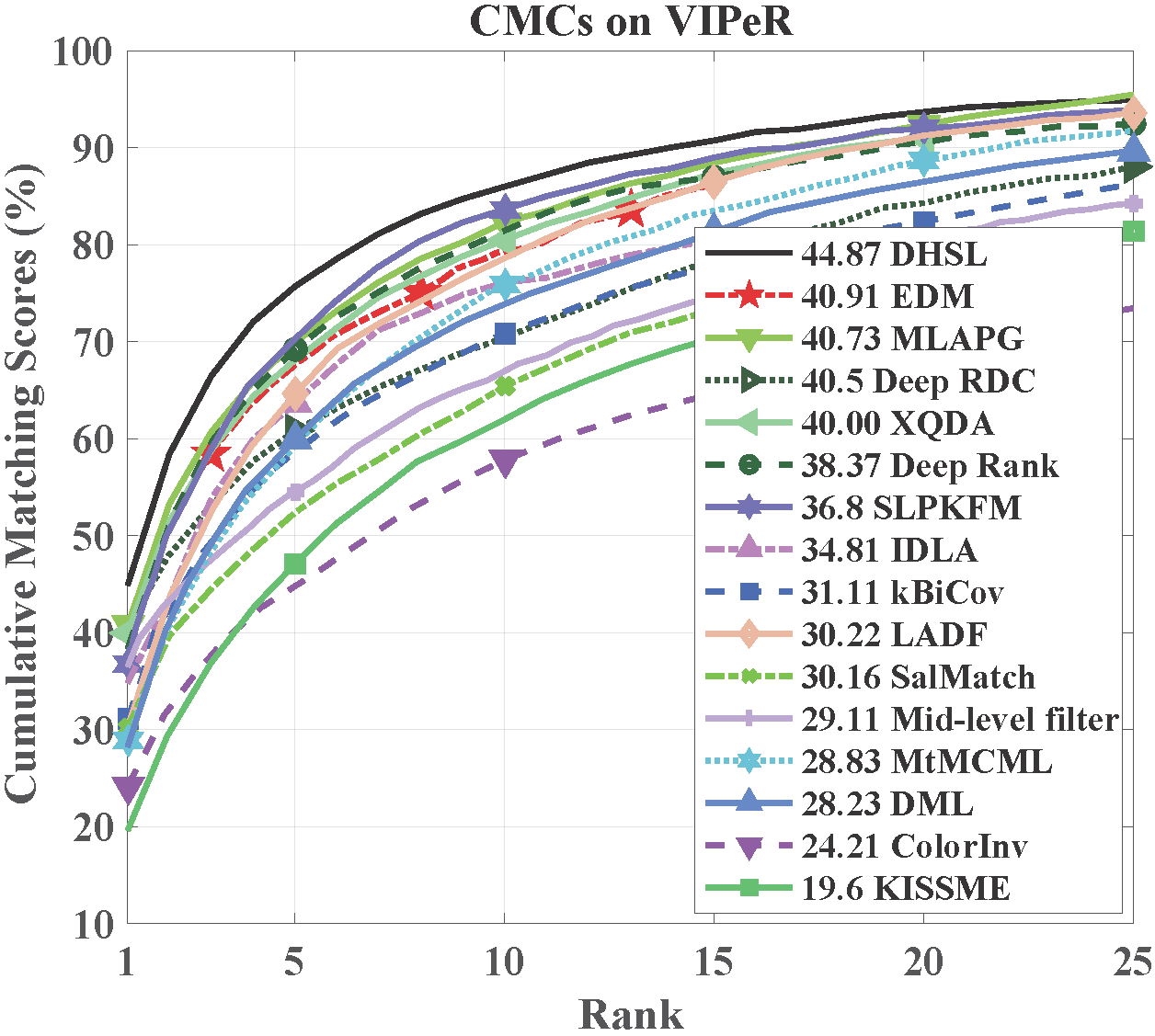}
    \caption{CMC curves and rank-1 identification rates on VIPeR \cite{viper} (gallery: 316 individuals).}
    \label{fig:viper_result}
\end{figure}


\begin{table}[tp]
\centering
\caption{Performance comparison between our proposed DHSL method and multiple state-of-the-art methods on VIPeR \cite{viper} (gallery: 316 individuals). {\color{red}Red}, {\color{green}green} and {\color{blue}blue} colors represent the {\color{red}1st}, {\color{green}2nd} and {\color{blue}3rd} best results, respectively.}\label{tab:viper}
\setlength{\tabcolsep}{1.2pt}
\begin{tabular}{c|c|c|c|c}
\hlinewd{1.5pt}
   Method & Rank=1 (\%) & Rank=10 (\%) & Rank=20 (\%) & Reference \\
\hlinewd{0.8pt}
  DHSL             & {\color{red}{\textbf{44.87}}} & {\color{red}{\textbf{86.01}}}  &  {\color{red}{\textbf{93.70}}}   &   proposed      \\
\hline
  MTL-LORAE         &{\color{green}{42.30}} &{\color{blue}{81.60}}   & {\color{blue}{89.60}}    & 2015 ICCV \cite{mtllr} \\
\hline
  EDM               &\color{blue}{40.91}   & N/A  & N/A  & 2016 ECCV \cite{edm}\\
\hline
  MLAPG             &{{40.73}}    &{\color{green}{82.34}}   & {\color{green}{92.37}}    & 2015 ICCV \cite{psd}\\
\hline
  Deep RDC          &40.5  &70.4     &84.4      & 2015 PR \cite{deepfeature}\\
\hline
  XQDA              &40.00 &80.51    & 91.08    & 2015 CVPR \cite{lomo}\\
\hline
  FT-JSTL+DGD       &38.6  &N/A    & N/A         &2016 CVPR \cite{dgd}\\
\hline
  Deep Rank      &38.37 &81.33  &90.43       &2016 TIP \cite{deeprank}\\
\hline
  SSDAL             &37.9  &75.6     & 85.4     & 2016 ECCV \cite{ssdal}\\
\hline
  SCNCD             &37.8  &81.2     &90.4     & 2014 ECCV \cite{colorname}\\
\hline
  GSCNN             &37.8  &77.4     &N/A       & 2016 ECCV \cite{gscnn} \\
\hline
  SLPKFM            &36.8  &83.7     & 91.7     & 2015 CVPR \cite{kfm} \\
\hline
  IDLA              &34.81 &76.12    &N/A      & 2015 CVPR \cite{improvedcnn}\\
\hline
  kBiCov            &31.11 &70.71    &82.44    & 2014 IVC \cite{kBiCov}\\
\hline
  LADF              &30.22 &78.82    & 90.44    & 2013 CVPR \cite{ladf} \\
\hline
  SalMatch          &30.16 &65.54    &79.15    & 2013 ICCV \cite{salienceiccv}\\
\hline
Mid-level filter    &29.11 &65.95    &79.87    & 2014 CVPR \cite{midlevelfeature}\\
\hline
  MtMCML            &28.83 &75.82    & 88.51    &  2014 TIP \cite{mtml} \\
\hline
  DML               &28.23 &73.45    &86.39     & 2014 ICPR \cite{dml} \\
\hline
  ColorInv          &24.21 &57.09    &69.65    & 2013 TPAMI \cite{color}\\
\hline
  KISSME            &19.6 &62.2    &77.0     & 2012 CVPR \cite{kissme}\\
\hlinewd{1.5pt}
\end{tabular}
\end{table}
\subsubsection{\textbf{Result on VIPeR}}
 Fig. \ref{fig:viper_result} and Table \ref{tab:viper} show the performance comparisons of the proposed DHSL method and the state-of-the-art person Re-ID methods on VIPeR \cite{viper}. One can see that DHSL outperforms CNN based person Re-ID methods (i.e. EDM \cite{edm}, Deep RDC \cite{deepfeature}, FT+JSTL+DGD \cite{dgd}, Deep Rank \cite{deeprank}, SSDAL \cite{ssdal}, GSCNN \cite{gscnn}, IDLA \cite{improvedcnn} and DML \cite{dml}) and metric learning based person Re-ID methods (i.e., MTL-LORAE \cite{mtllr}, MLAPG \cite{psd} and XQDA \cite{lomo}). For example, DHSL improves the rank-1 identification rate by 3.96\%, 4.37\% and 6.27\% over EDM \cite{edm}, Deep RDC \cite{deepfeature} and FT+JSTL+DGD \cite{dgd}, respectively. DHSL beats MTL-LORAE \cite{mtllr}, MLAPG \cite{psd} and XQDA \cite{lomo} by 2.57\%, 4.14\% and 4.87\%, respectively, at rank 1. Moreover, the training of DHSL is much simpler than FT+JSTL+DGD \cite{dgd} and SSDAL \cite{ssdal}. Because DHSL does not require a large database to pre-train the deep CNN, compared with FT+JSTL+DGD \cite{dgd}. DHSL also does not need to use additional human attributes to pre-train the deep CNN as that in SSDAL \cite{ssdal}.

\subsubsection{\textbf{Result on CUHK03}}
\begin{figure*}[tp]
\subfigure[]{
\begin{minipage}[t]{0.33\linewidth}
\centering
\includegraphics[width=0.9\linewidth]{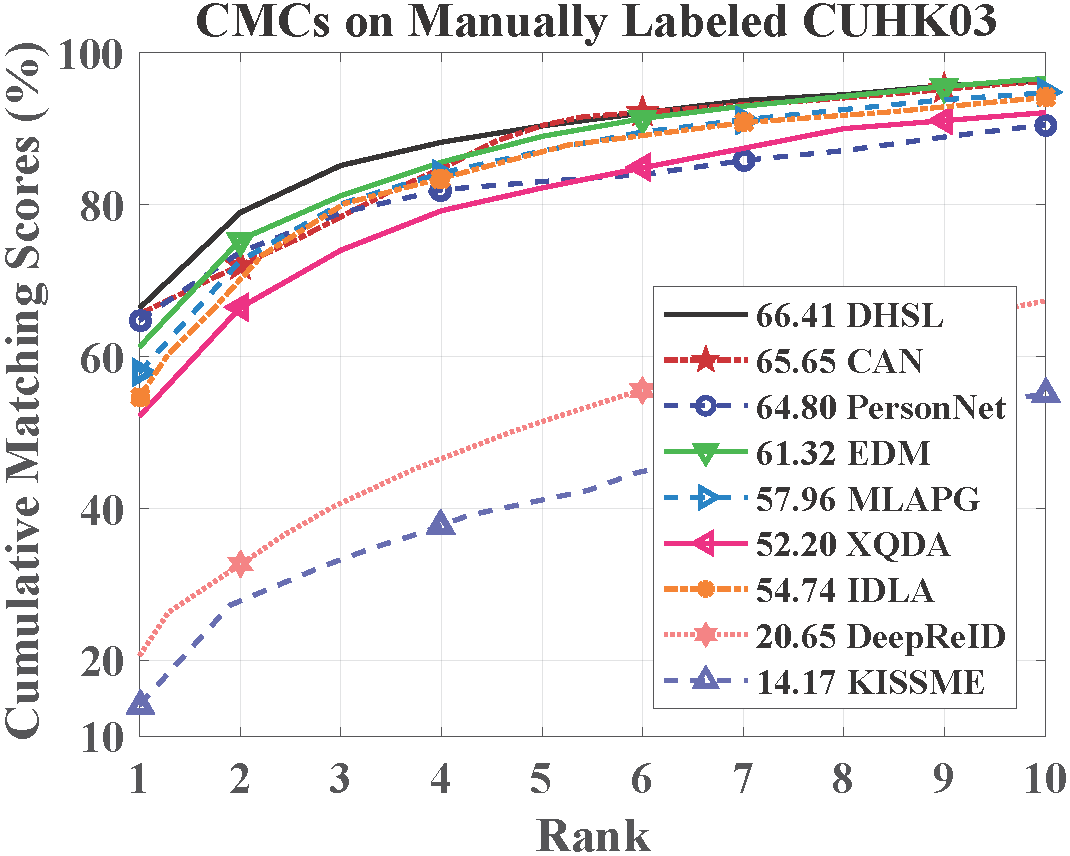}
\label{fig:cuhk03:a}
\end{minipage}
}%
\subfigure[]{
\begin{minipage}[t]{0.33\linewidth}
\centering
\includegraphics[width=0.9\linewidth]{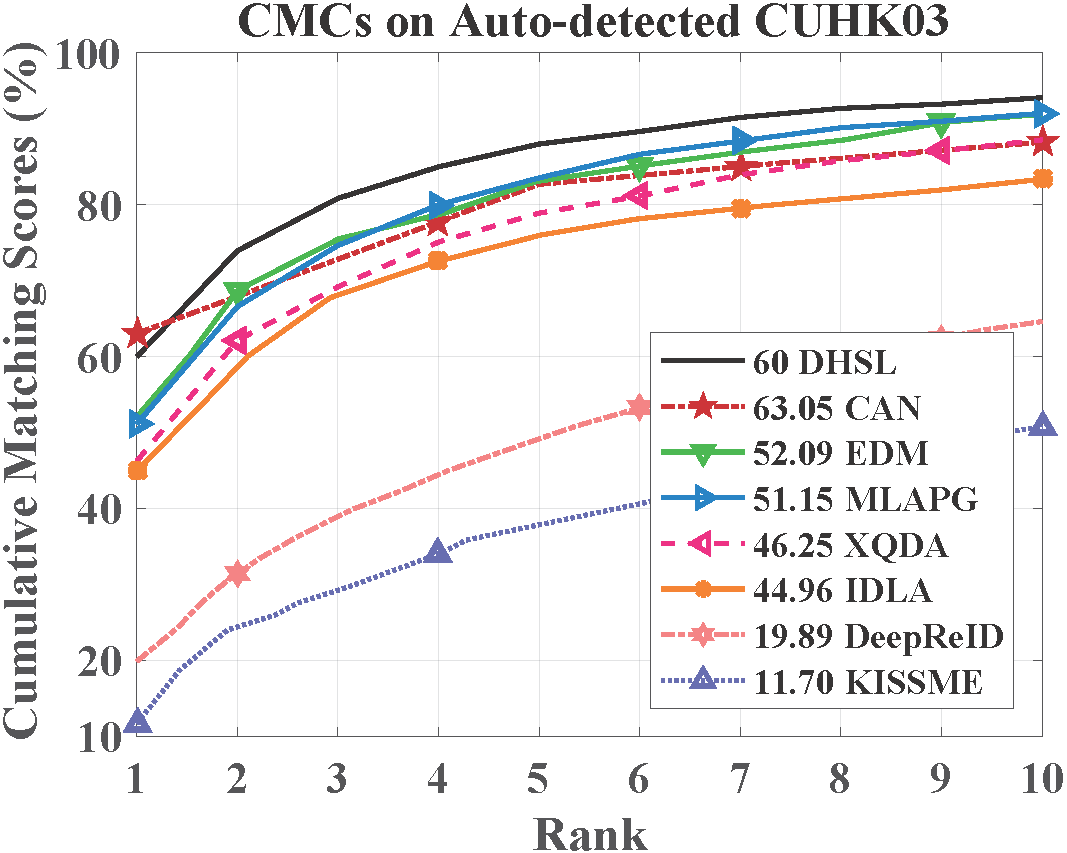}
\label{fig:cuhk03:b}
\end{minipage}
}
\subfigure[]{
\begin{minipage}[t]{0.33\linewidth}
\centering
\includegraphics[width=0.9\linewidth]{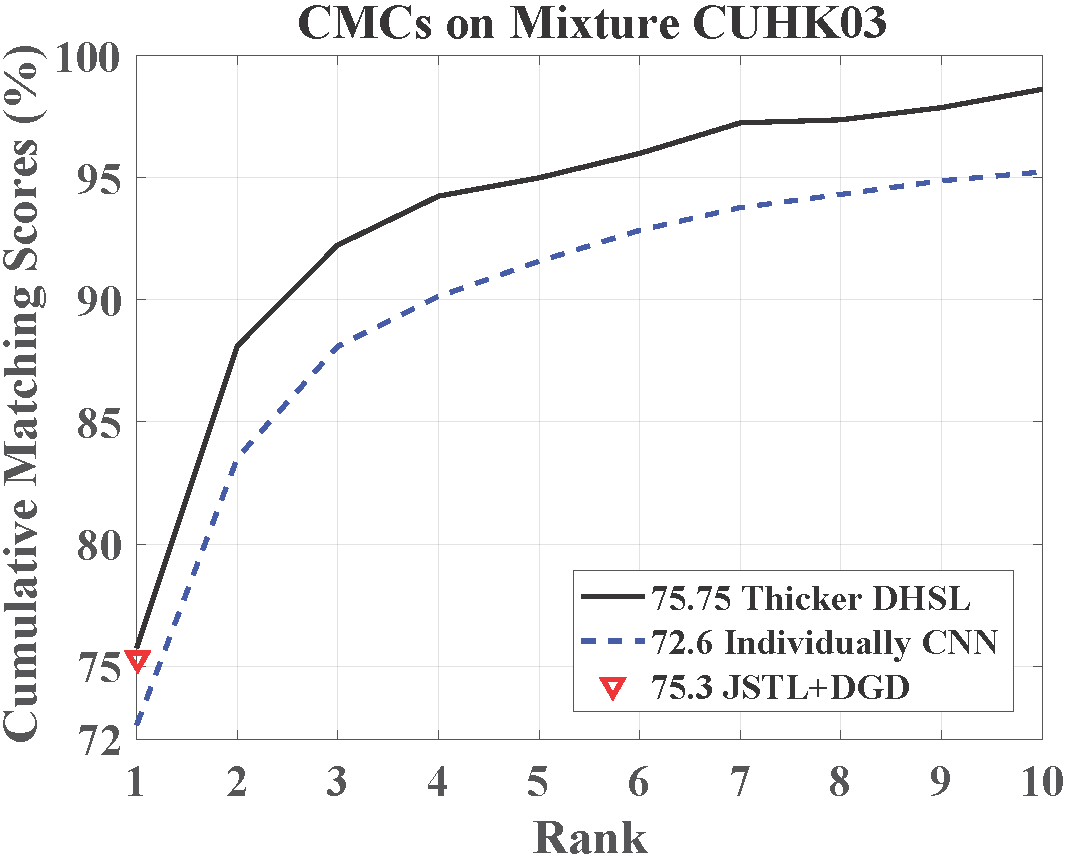}
\label{fig:cuhk03:c}
\end{minipage}
}
\caption{CMC curves and rank-1 identification rates on CUHK03 \cite{cuhk03} (gallery: 100 individuals). (a) Manually labeled setting. (b) Auto-detected setting. (c) Mixture setting, which mixes the manually labeled and auto-detected settings together for obtaining a large training set. } \label{fig:chhk03}
\end{figure*}

 Fig. \ref{fig:chhk03} shows the performance comparison between the proposed DHSL method and the state-of-the-art person Re-ID methods CUHK03 \cite{cuhk03}.

 As shown in Fig. \ref{fig:chhk03}(a), DHSL beats the CNN based person Re-ID methods (i.e. CAN \cite{can}, PersonNet \cite{personnet}, EDM \cite{edm}, IDLA \cite{improvedcnn}, DeepReID \cite{cuhk03}) and the metric learning based person Re-ID methods (i.e. MLAPG \cite{psd}, XQDA \cite{lomo} and KISSME \cite{kissme}) on the manually labeled CUHK03 setting. As shown in Fig. \ref{fig:chhk03}(b), on the auto-detected CUHK03 setting, the similar comparison result is obtained, although the rank-1 identification rate of DHSL is bit lower than that of the CAN \cite{can} method.

 In \cite{dgd}, both the domain individually CNN and domains jointed in single-task learning with domain guided dropout (JSTL+DGD) person Re-ID models are trained on a mixture setting. The mixture setting are constructed by mixing the manually labeled and auto-detected settings together. Since the mixture setting is large enough, we trained a thicker DHSL model. The number of channel in each layer in the thicker DHSL model are twice of that in the original DHSL model. For example, the C1, C2 and C3 layers in the thicker DHSL model hold 64, 128 and 256 channels, respectively. The thickest layers in individually CNN and JSTL+DGD models hold 1536 channels, thus the thicker DHSL model is still thinner than individually CNN and JSTL+DGD. As shown in Fig. \ref{fig:chhk03} (c), the thicker DHSL obviously outperforms the individually CNN \cite{dgd}. Moreover, thicker DHSL obtains a bit higher rank 1 identification rate than JSTL+DGD.


 Based on the extensive experiments on either small datasets QMUL GRID, VIPeR or large dataset CUHK03, one can find that the proposed method outperforms both the-state-of-art CNN and metric learning based person Re-ID methods. These results validate the effectiveness of the proposed DHSL method.

\subsection{Analysis of the Proposed DHSL Method}
 Recall that the proposed DHSL method consists of two learning modules. In this subsection, we further make a comprehensive performance analysis to show the effectiveness of each module in the proposed DHSL method, as follows.
\begin{figure}[t]
    \centering
    \includegraphics[width=.75\linewidth]{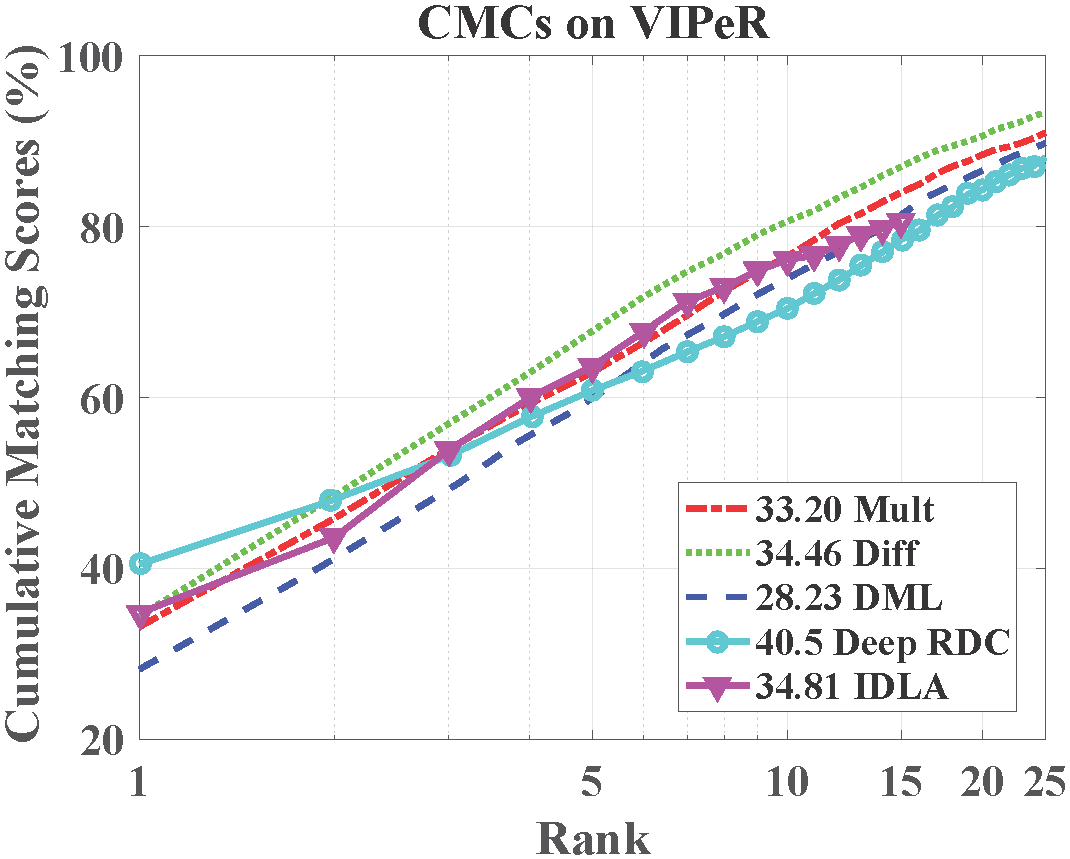}
    \caption{CMC curves and rank-1 identification rates on VIPeR \cite{viper} (gallery: 316 individuals) resulted from three state-of-the-art CNN based person Re-ID methods and the proposed DHSL method under the element-wise absolute difference (Diff) and the element-wise multiplication (Mult) learning configurations, respectively.}
    \label{fig:viper_role_of_cnn}
\end{figure}
\subsubsection{\textbf{Role of the proposed CNN Feature Learning Module}}
 To validate the effectiveness of the proposed CNN feature learning module, experiments are conducted under either the element-wise absolute difference (Diff) or the element-wise multiplication (Mult) configurations  rather than the proposed hybrid similarity function. By assigning different values to $W_d$ and $W_m$ in Eq. (\ref{equ:hybrid}), we can obtain the performance by considering only Diff ($W_m \equiv 0$) or Mult ($W_d \equiv 0$). The results are shown in Fig. \ref{fig:viper_role_of_cnn}. Note that with the Diff or Mult configuration, the trained CNN based person Re-ID model is similar to that used in the state-of-the-art CNN based person Re-ID models (i.e., IDLA \cite{improvedcnn}, Deep RDC \cite{deepfeature} and DML \cite{dml}), since these methods train their person Re-ID models on a single feature space constructed by calculating differences or multiplications of feature pairs.

 From Fig. \ref{fig:viper_role_of_cnn}, one can see that the result by using the specially designed CNN under the Diff configuration is better than Deep RDC and IDML. Moreover, the result by using the specially designed CNN under the Mult configuration is also a bit better than that of DML. This study demonstrates the effectiveness of the proposed CNN feature learning module.

\begin{table}[t]
\centering
\caption{Top Ranked identification rates under different feature dimensions on QMUL GRID \cite{grid}. Dim represents the compressed feature dimension.}\label{tab:grid_hyvsmah}
\setlength{\tabcolsep}{1.8pt}
\begin{tabular}{c|c|c|c|c|c}
\hlinewd{1.5pt}
  Method       & Dim & \begin{tabular}{c}
                   Rank=1\\
                   (\%)  \\
                 \end{tabular}
               & \begin{tabular}{c}
                   Rank=10\\
                   (\%)  \\
                 \end{tabular}
               & \begin{tabular}{c}
                   Rank=20\\
                   (\%)  \\
                 \end{tabular}
               & \begin{tabular}{c}
                   Rank=30\\
                   (\%)  \\
                 \end{tabular} \\
\hlinewd{0.8pt}
  DHSL          &-&\textbf{21.20} &\textbf{54.24}    &\textbf{65.84}    &\textbf{71.28}   \\
\hlinewd{0.8pt}
  \multirow{5}{*}{CNNFeat+PCA+KISSME}
  & 16 & 12.24 & 41.04 & 54.96 & 63.68 \\
\cline{2-6}
  & 32 & 16.96 & 47.68 & 58.64 & 66.48 \\
\cline{2-6}
  & 64 & 17.20 & 44.32 & 54.80 & 61.60 \\
\cline{2-6}
  & 128 & 15.04 & 39.36 & 50.48 & 57.12 \\
\cline{2-6}
  & 256 & 17.12 & 40.96 & 49.68 & 56.40 \\
\hlinewd{0.8pt}
   \multirow{5}{*}{CNNFeat+PCA+ITML}
  & 16 & 5.84 & 25.44 & 34.00 & 38.96 \\
\cline{2-6}
  & 32 & 7.92 & 29.20 & 40.16 & 46.80 \\
\cline{2-6}
  & 64 & 9.76 & 35.92 & 47.44 & 54.72 \\
\cline{2-6}
  & 128 & 13.60 & 42.08 & 52.96 & 61.60 \\
\cline{2-6}
  & 256 & 14.32 & 44.40 & 55.68 & 63.60 \\
\hlinewd{0.8pt}
   \multirow{5}{*}{CNNFeat+PCA+MLAPG}
  & 16 & 9.76 & 37.52 & 50.24 & 59.12 \\
\cline{2-6}
  & 32 & 15.76 & 43.12 & 56.64 & 64.24 \\
\cline{2-6}
  & 64 & 14.96 & 44.08 & 57.36 & 66.08 \\
\cline{2-6}
  & 128 & 15.60 & 42.80 & 57.04 & 65.28 \\
\cline{2-6}
  & 256 & 15.36 & 42.40 & 56.24 & 64.96 \\
\hlinewd{0.8pt}
   \multirow{5}{*}{CNNFeat+XQDA}
  & 16 & 12.88 & 41.36 & 54.48 & 63.20 \\
\cline{2-6}
  & 32 & 13.52 & 43.28 & 55.76 & 62.96 \\
\cline{2-6}
  & 64 & 16.08 & 41.68 & 54.32 & 61.12 \\
\cline{2-6}
  & 128 & 15.60 & 39.84 & 49.76 & 55.04 \\
\cline{2-6}
  & 256 & 15.60 & 39.76 & 49.76 & 55.04 \\
\hlinewd{0.8pt}
    \multirow{5}{*}{CNNStru+FC+Mah}
  & 16 & 2.72 & 22.56 & 37.76 & 47.36 \\
\cline{2-6}
  & 32 & 3.36 & 24.48 & 38.64 & 47.52 \\
\cline{2-6}
  & 64 & 3.92 & 26.00 & 39.04 & 48.80 \\
\cline{2-6}
  & 128 & 4.16 & 25.84 & 38.56 & 49.04 \\
\cline{2-6}
  & 256 & 4.00 & 26.48 & 40.00 & 47.60 \\
\hlinewd{1.5pt}
\end{tabular}
\end{table}
\begin{table}[t]
\centering
\caption{Top Ranked identification rates under different feature dimensions on VIPeR \cite{viper}. Dim represents the compressed feature dimension.}\label{tab:viper_hyvsmah}
\setlength{\tabcolsep}{1.8pt}
\begin{tabular}{c|c|c|c|c|c}
\hlinewd{1.5pt}
  Method       & Dim & \begin{tabular}{c}
                   Rank=1\\
                   (\%)  \\
                 \end{tabular}
               & \begin{tabular}{c}
                   Rank=10\\
                   (\%)  \\
                 \end{tabular}
               & \begin{tabular}{c}
                   Rank=20\\
                   (\%)  \\
                 \end{tabular}
               & \begin{tabular}{c}
                   Rank=30\\
                   (\%)  \\
                 \end{tabular} \\
\hlinewd{0.8pt}
  DHSL          &-&\textbf{44.87} &\textbf{86.01}    &\textbf{93.70}    &\textbf{95.89}   \\
\hlinewd{0.8pt}
  \multirow{5}{*}{CNNFeat+PCA+KISSME}
    &16& 19.24 & 63.73 & 80.03 & 88.20 \\
\cline{2-6}
    &32& 26.68 & 72.59 & 85.82 & 91.87 \\
\cline{2-6}
    &64& 32.25 & 74.56 & 86.55 & 91.74 \\
\cline{2-6}
    &128& 30.76 & 71.08 & 82.94 & 89.27  \\
\cline{2-6}
    &256& 27.37 & 65.09 & 77.59 & 83.26 \\
\hlinewd{0.8pt}
   \multirow{5}{*}{CNNFeat+PCA+ITML}
  & 16 & 5.82 & 22.34 & 30.89 & 37.50 \\
\cline{2-6}
  & 32 & 11.93 & 38.35 & 49.08 & 55.79 \\
\cline{2-6}
  & 64 & 20.19 & 58.54 & 71.87 & 79.30 \\
\cline{2-6}
  & 128 & 17.15 & 51.08 & 63.86 & 70.76 \\
\cline{2-6}
  & 256 & 25.57 & 67.53 & 80.70 & 87.18 \\
\hlinewd{0.8pt}
   \multirow{5}{*}{CNNFeat+PCA+MLAPG}
  & 16 & 16.39 & 60.28 & 77.75 & 85.98 \\
\cline{2-6}
  & 32 & 23.07 & 69.40 & 83.29 & 89.30 \\
\cline{2-6}
  & 64 & 29.18 & 73.13 & 86.17 & 91.33 \\
\cline{2-6}
  & 128 & 28.99 & 73.48 & 85.95 & 91.77 \\
\cline{2-6}
  & 256 & 28.10 & 72.69 & 85.66 & 91.33 \\
\hlinewd{0.8pt}
   \multirow{5}{*}{CNNFeat+XQDA}
  &16 & 23.92 & 68.26 & 82.72 & 89.78 \\
\cline{2-6}
  &32 & 28.39 & 70.09 & 82.41 & 89.21 \\
\cline{2-6}
  &64 & 28.67 & 67.34 & 79.75 & 86.27 \\
\cline{2-6}
  &128 & 26.33 & 61.42 & 74.43 & 81.11 \\
\cline{2-6}
  &256 & 23.51 & 55.32 & 67.31 & 74.21 \\
\hlinewd{0.8pt}
    \multirow{5}{*}{CNNStru+FC+Mah}
  & 16 & 15.16 & 58.89 & 75.85 & 83.96 \\
\cline{2-6}
  & 32 & 16.90 & 63.86 & 79.49 & 86.93 \\
\cline{2-6}
  & 64 & 17.34 & 64.65 & 79.59 & 87.03 \\
\cline{2-6}
  & 128 & 18.64 & 63.83 & 79.08 & 86.90 \\
\cline{2-6}
  & 256 & 17.91 & 64.78 & 79.24 & 86.33 \\
\hlinewd{1.5pt}
\end{tabular}
\end{table}

\begin{figure}[t]
    \centering
    \includegraphics[width=0.8\linewidth]{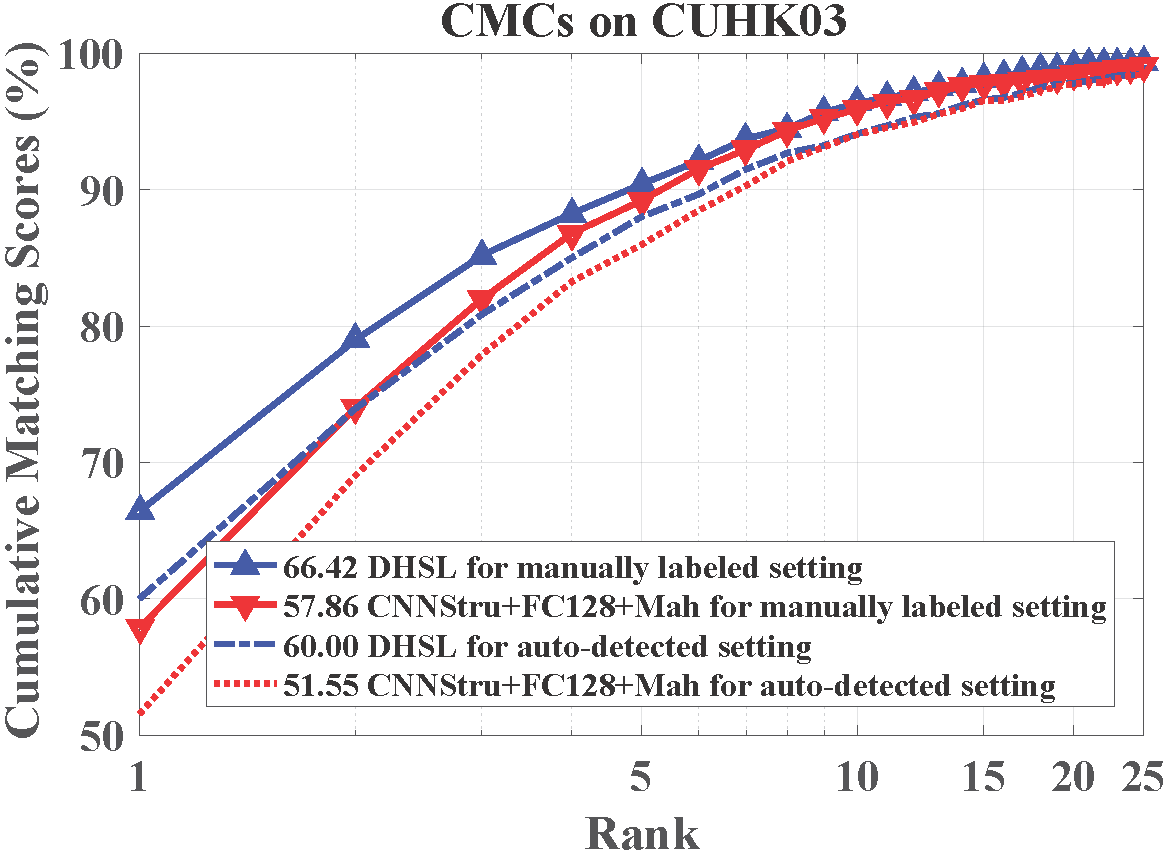}
    \caption{CMC curves and rank-1 identification rates on CUHK03 \cite{cuhk03} (gallery: 100 individuals). }
    \label{fig:cuhk03_mah}
\end{figure}

\subsubsection{\textbf{Role of Proposed Hybrid Similarity Module}}%
 In addition to the specially designed CNN feature learning module, extensive experiments are also conducted to evaluate the effectiveness of the proposed hybrid similarity learning module. The experiments consist of the following two parts: (1) the performance comparison between the proposed hybrid similarity function and the Mahalanobis distance based similarity metrics learned by the state-of-art metric learning methods, i.e., KISSME \cite{kissme}, ITML \cite{itml}, MLAPG \cite{psd}, and XQDA \cite{lomo}, using the same CNN feature (CNNFeat) that used in the proposed hybrid similarity function; (2) the performance comparison to the similarity metric by incorporating the Mahalanobis distance function into the proposed specially designed CNN feature learning module. Moreover, since the dimension of the feature (i.e., the output of the A1 layer in Fig. \ref{fig:framework}) extracted by the proposed CNN feature learning module is $16 \times 128=2048$ and the number of parameters in the Mahalanobis matrix is square of the feature dimension, a feature compression operation is necessary before the above-mentioned Mahalannobis distance based similarity metric learning. For that, in experimental part (1), for the methods, KISSME \cite{kissme}, ITML \cite{itml} and MLAPG \cite{psd}, the PCA is used to compress the feature dimension, while the XQDA \cite{lomo} can automatically realize the feature dimension compression. Hence, these methods are denoted as CNNFeat+PCA+KISSME, CNNFeat+PCA+ITML, CNNFeat+PCA+MLPAG, and CNNFeat+XQDA, respectively. In experimental part (2), the bottom CNN feature learning module (see Fig. \ref{fig:framework}, C1$\rightarrow$...$\rightarrow$A1) is kept the same with that is used for the proposed hybrid similarity function. Differently, an additional full connection (FC) layer is integrated after the A1 layer of the proposed CNN feature learning module to realize the feature compression. Hence, this method is denoted as CNNStru+FC+Mah. Moreover, considering that the feature compression degree has an important influence on the performance, the experiments are performed under different compressed feature dimensions. The corresponding results are shown in Tables \ref{tab:grid_hyvsmah}, \ref{tab:viper_hyvsmah} and Fig. \ref{fig:cuhk03_mah}.

 As shown in Tables \ref{tab:grid_hyvsmah} and \ref{tab:viper_hyvsmah}, the proposed method obtains higher recognition rates than CNNFeat+PCA+KISSME, CNNFeat+PCA+ITML, CNNFeat+PCA+MLAPG, and CNNFeat+XQDA both on QMUL GRID and VIPeR datasets. This is because these methods independently optimize the feature learning and metric learning, while the proposed hybrid similarity function is able to jointly optimize the feature learning and metric learning. More specifically, for CNNFeat+PCA+KISSME, CNNFeat+PCA+ITML and CNNFeat+PCA+MLAPG, the feature compression via PCA does not consider the metric learning in general. For CNNFeat+XQDA, although XQDA can find an optimal subspace for metric learning, it still requires a large number of parameters in the feature compression, which is prone to over-fitting on a small dataset. On the contrary, the proposed method does not require a feature compression operation. Consequently, the proposed hybrid similarity function achieves a superior performance.

 In addition, the proposed hybrid similarity function also obtains obvious improvement both on small (i.e., QMUL GRID and VIPeR) and large (i.e., CHUK03) datasets, compared with CNNStru+FC+Mah, as shown in Tables \ref{tab:grid_hyvsmah} and \ref{tab:viper_hyvsmah}, and Fig. \ref{fig:cuhk03_mah}. Note that for the large dataset CUHK03, only the compressed feature dimension 128 is learned and denoted as CNNStru+FC128+Mah, as it has better result on QMUL and VIPeR datasets. The superior performance achieved by the proposed DHSL method further indicates that the proposed hybrid similarity function is more suitable to be integrated with a CNN for person Re-ID than the Mahalannobis distance function, since it can save a lot of parameter in the similarity metric learning so as to relieve the over-fitting problem.

\subsubsection{\textbf{Role of the Element-wise Absolute Difference and Multiplication Complementary Behavior}}
\begin{figure}[t]
    \centering
    \includegraphics[width=1.0\linewidth]{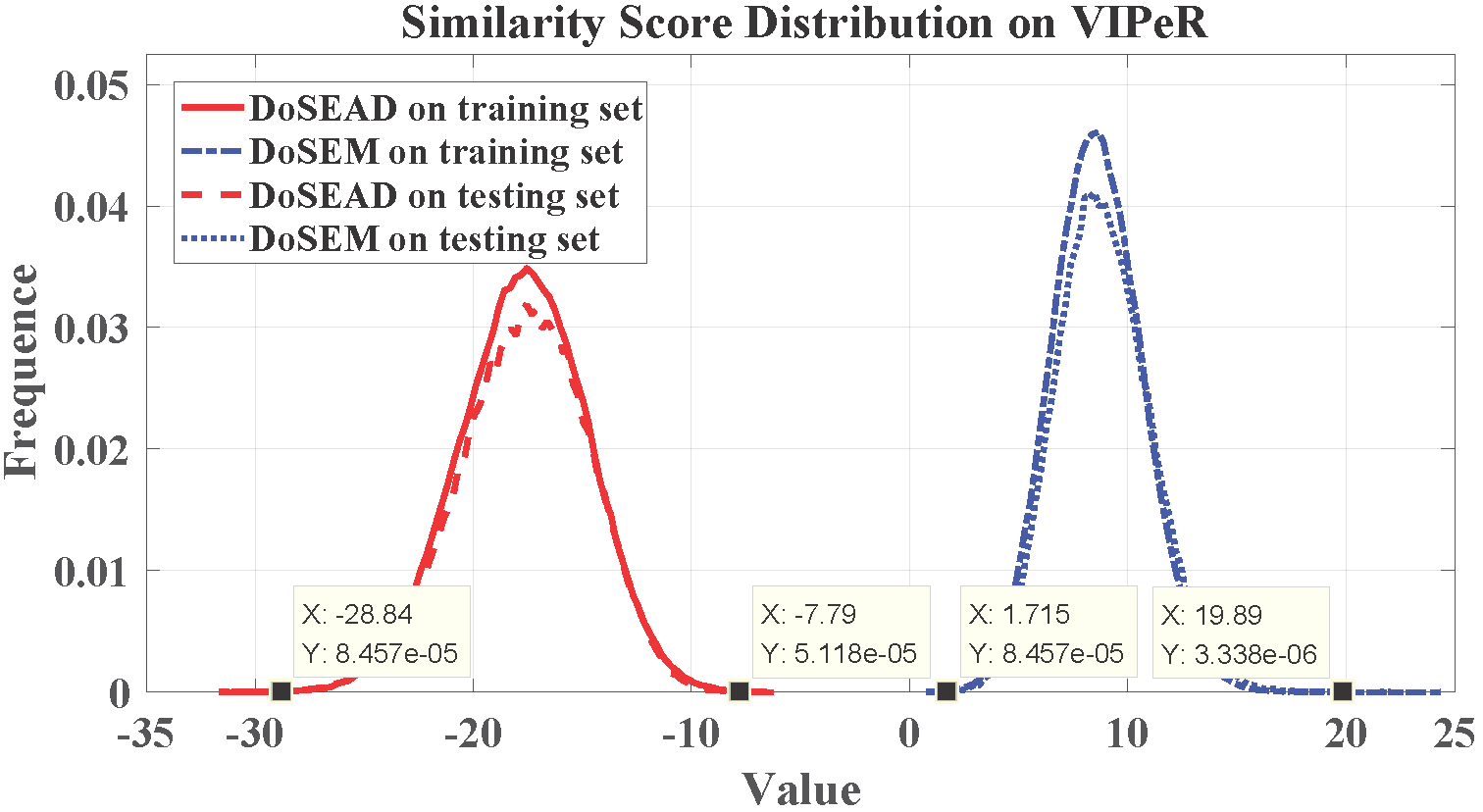}
    \caption{Similarity score distributions on VIPeR \cite{viper}. DoSEAD represents the Distribution of similarity Scores calculated by projecting Element-wise Absolute Differences with $W_d$ in Eq. (\ref{equ:hybrid}). DoSEM represents the Distribution of similarity Scores calculated by projecting Element-wise Multiplications with $W_m$ in Eq. (\ref{equ:hybrid}).}
    \label{fig:viper_scale_analy}
\end{figure}

 First, we validate that it is able to learn a hybrid similarity function by projecting element-wise absolute differences and multiplications into similarity scores simultaneously. As shown in the Fig. \ref{fig:framework}, there are three parameter shared batch normalization layers in each feature extraction branch of the proposed CNN feature learning module, which could limit the scale of output features. Moreover, the scales of these two features are further adjusted by the learned parameters ($W_d$ and $W_m$ in Eq. (\ref{equ:hybrid})). The Distribution of similarity Scores calculated by projecting Element-wise Absolute Differences (DoSEAD) with $W_d$) and the Distribution of similarity Scores calculated by projecting Element-wise Multiplications (DoSDM) with $W_m$) are evaluated, as shown in Fig. \ref{fig:viper_scale_analy}. One can see that both on the training and testing setting of VIPeR \cite{viper}, the rough ranges of DoSEAD and DoSEM are [-28.84, -7.79] and [1.715, 19.89], respectively. This study indicates that the element-wise absolute difference and multiplication metrics have no large scale difference, therefore it is able to learn a hybrid similarity function that considers element-wise absolute difference and multiplication simultaneously.

\begin{figure}[t]
    \centering
    \includegraphics[width=.75\linewidth]{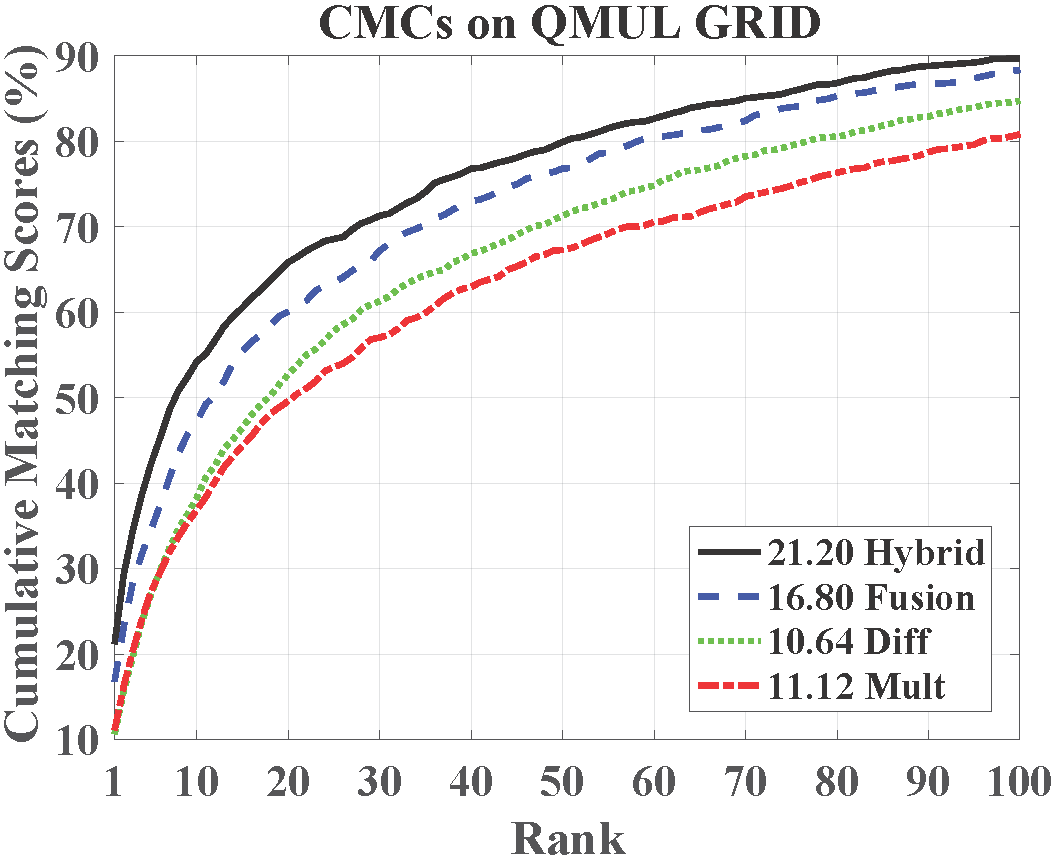}
    \caption{CMC curves and rank-1 identification rates on QMUL GRID \cite{grid} (gallery: 125 individuals+775 background images) resulted from the proposed DHSL method under the element-wise absolute difference (Diff), the element-wise multiplication (Mult), the simple score fusion (Fusion) of Diff and Mult, and the hybrid similarity learning configurations, respectively.}
    \label{fig:grid_comple}
\end{figure}

\begin{figure}[t]
    \centering
    \includegraphics[width=.75\linewidth]{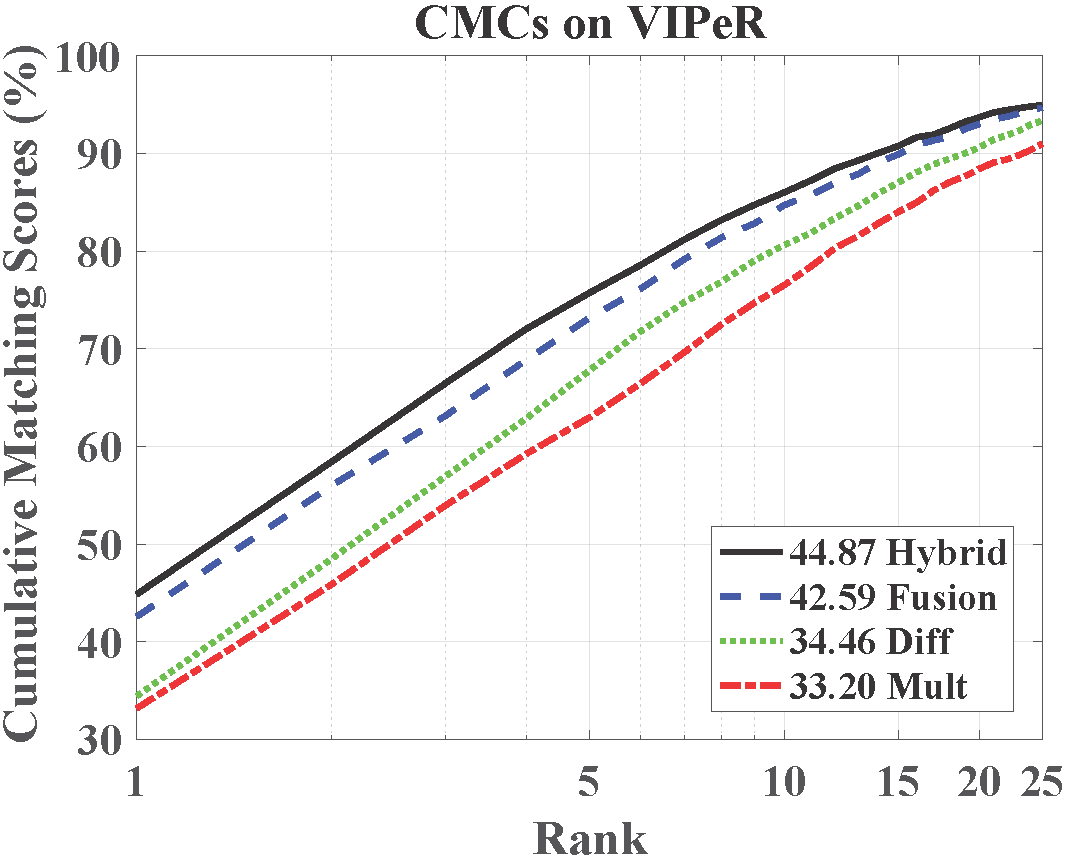}
    \caption{CMC curves and rank-1 identification rates on VIPeR \cite{viper} (gallery: 316 individuals) resulted from the proposed DHSL method under the element-wise absolute difference (Diff), the element-wise multiplication (Mult), the simple score fusion (Fusion) of Diff and Mult, and the hybrid similarity learning configurations, respectively.}
    \label{fig:viper_comple}
\end{figure}

 Second, we validate the proposed hybrid similarity function simultaneously learns on element-wise absolute differences and multiplications is helpful for improving person Re-ID performance. We evaluate the proposed DHSL method under the element-wise absolute difference (Diff), the element-wise multiplication (Mult), the simple score fusion of Diff and Mult (Fusion), and the proposed hybrid similarity learning (Hybrid) configurations, respectively. The fusion configuration is to independently train Diff and Mult person Re-ID models under the Diff and Mult configurations and then simply summarize the similarity scores from the Diff and Mult person Re-ID models as the final similarity score.

 It can be observed from the results shown in Figs. \ref{fig:grid_comple} and \ref{fig:viper_comple} that the CMC resulted from the Fusion method is obvious better than that resulted from either the Diff or Mult method. This indicates the element-wise absolute difference and multiplication play an effective complementary role to each other for improving the person Re-ID performance. Moreover, it can be further found that the proposed hybrid similarity method (Hybrid) outperforms the Fusion method. This is due to the fact that the proposed deep hybrid similarity learning method is able to maximize the complementary effectiveness between Diff and Mult.


\begin{table}[t]
\centering
\caption{Top Ranked identification rates under different training dataset scales on QMUL GRID \cite{grid}.}\label{tab:grid_traingscale}
\setlength{\tabcolsep}{1.8pt}
\begin{tabular}{c|c|c|c|c}
\hlinewd{1.5pt}
     \begin{tabular}{c}
                                   Training Scale\\
                                   (Individuals) \\
     \end{tabular}
               & Rank=1 (\%) & Rank=10 (\%) & Rank=20 (\%) & Rank=30 (\%) \\
\hlinewd{0.8pt}
50	&14.48	&	42.72	&	54.96	&	63.36\\
\hline
75	&18.48	&	46.64	&	58.64	&	66.48\\
\hline
100	&\textbf{21.20}	&	51.44	&	63.84	&	70.40\\
\hline
125	&\textbf{21.20}	&	\textbf{54.24}	&	\textbf{65.84}	&	\textbf{71.28}\\
\hlinewd{1.5pt}
\end{tabular}
\end{table}
\begin{table}[htp]
\centering
\caption{Top Ranked identification rates under different training dataset scales on VIPeR \cite{viper}.}\label{tab:viper_traingscale}
\setlength{\tabcolsep}{1.8pt}
\begin{tabular}{c|c|c|c|c}
\hlinewd{1.5pt}
     \begin{tabular}{c}
                                   Training Scale\\
                                   (Individuals) \\
     \end{tabular}
               & Rank=1 (\%) & Rank=10 (\%) & Rank=20 (\%) & Rank=30 (\%) \\
\hlinewd{0.8pt}
  100          &30.79 &72.78    &83.92    &88.70   \\
\hline
  150          &34.72 &78.86    &88.51    &92.91   \\
\hline
  200          &41.61 &81.90    &91.23    &94.56   \\
\hline
  250          &41.77 &83.54    &92.25    &95.35   \\
\hline
  300          &44.08 &84.97    &93.01    &95.70   \\
\hline
  316          &\textbf{44.87} &\textbf{86.01 }   &\textbf{93.70}    &\textbf{95.89}    \\
\hlinewd{1.5pt}
\end{tabular}
\end{table}

\subsubsection{\textbf{Role of Training Dataset Scale}}

 In this experiment, the number of training person individuals is changed, and the number of testing person image pairs is kept the same (i.e., 125 testing person individuals on QMUL GRID \cite{grid}, 316 testing person individuals on VIPeR \cite{viper}).

 From the results shown in Tables \ref{tab:grid_traingscale} and \ref{tab:viper_traingscale}, one can observe that the performance of the proposed DHSL method is increased with the training dataset scale on both QMUL GRID \cite{grid} and VIPeR \cite{viper} datasets. Moreover, from the results on QMUL GRID in Tables \ref{tab:grid} and \ref{tab:grid_traingscale}, it can be seen that the proposed DHSL method trained by only using 75 training person individuals obtains a better performance than MLAPG \cite{psd}, SLPKFM \cite{kfm}, XQDA \cite{lomo} and MRank-RankSVM \cite{mrk}. Similarly, from the results on VIPeR in Tables \ref{tab:viper} and \ref{tab:viper_traingscale}, it can be found that the proposed DHSL method trained by only using 200 training person individuals is comparable to MTL-LORAE \cite{mtllr}, MLAPG \cite{psd}, XQDA \cite{lomo} and Deep RDC \cite{deepfeature}, and is better than the rest methods, such as LADF \cite{ladf}, MtMCML \cite{mtml}, KISSME \cite{kissme}, IDLA \cite{improvedcnn} and DML \cite{dml}, and so on. These results illustrate that the proposed DHSL method is less dependent on the training data scale. This is because the proposed DHSL method has a reasonable number of parameters.

\subsubsection{\textbf{Running Time Analysis}}

\begin{table}[tp]
\centering
\caption{Running time comparison of different methods. FET and SCT represent feature extraction time per image and similarity calculation time per image pair, respectively.}\label{tab:speed}
\setlength{\tabcolsep}{0.8pt}
\begin{tabular}{c|c|c|c|c|c}
\hlinewd{1.5pt}
\multirow{2}{*}{Method} & \multirow{2}{*}{Mex} & \multicolumn{2}{c|}{\begin{tabular}{c}
           CPU: \\
           I7-6820HQ @2.7 Hz \\
         \end{tabular}} & \multicolumn{2}{c}{\begin{tabular}{c}
           GPU: \\
           NVIDIA Quadro M1000M \\
         \end{tabular}} \\
\cline{3-6}
       & & \begin{tabular}{c}
           FET \\
           (msec/image) \\
         \end{tabular}
       & \begin{tabular}{c}
           SCT \\
           (msec/pair) \\
         \end{tabular}
       & \begin{tabular}{c}
           FET \\
           (msec/pair)\\
         \end{tabular}
       & \begin{tabular}{c}
           SCT \\
           (msec/pair) \\
         \end{tabular}\\
\hlinewd{0.8pt}
  DHSL                &Yes        &\textbf{4.9781} & \textbf{0.0373} & \textbf{0.3640} & \textbf{0.0028}\\
\hline
  LOMO \cite{lomo}    &Yes        &10.1297  & N/A   & N/A    & N/A   \\
\hline
  ELF16  \cite{elf16} &No         &254.6033 & N/A   & N/A    & N/A   \\
\hlinewd{.8pt}
  Thicker DHSL        &Yes        &10.3066   &0.0749 &0.8462  &0.0053 \\
\hline
  JSTL+DGD \cite{dgd} &Yes        &111.0322  &   N/A & 5.9347 & N/A \\

\hlinewd{1.5pt}
\end{tabular}
\end{table}

  To validate the running time advantage of the proposed DHSL method, we evaluate the feature extraction time (FET) per image and similarity calculation time (SCT) per image pair. The software tools are Matconvnet \cite{matcovnnet}, CUDA 8.0, CUDNN V5.1, MATLAB 2016 and Visual Studio 2015. We re-implemented LOMO \footnote{\url{http://www.cbsr.ia.ac.cn/users/scliao/projects/lomo\_xqda/index.html}} \cite{lomo} and EFL16 \footnote{\url{http://isee.sysu.edu.cn/~chenyingcong/code/demo\_feat.zip}} \cite{elf16} feature extraction codes provided by their authors to evaluate FETs. Both for LOMO and EFL16, the FET excludes the feature compression running time, since the initial code does not provide feature compression code. The pretrained JSTL+DGD \footnote{\url{https://github.com/Cysu/dgd\_person\_reid}} \cite{dgd} model is implemented in the Caffe \cite{caffe} deep learning framework. The results are shown in Table \ref{tab:speed}.

  As shown in Table \ref{tab:speed}, the summation of FET and SCT resulted from DHSL is less than half of the FET of LOMO \cite{lomo}, using the same CPU setup. Moreover, the FETs of DHSL are only 4.48\% and 6.13\% of those of JSTL+DGD \cite{dgd} under the same CPU and GPU settings, respectively. Even for the thicker DHSL, its FETs are only 9.28\% and 14.26\% of those of JSTL+DGD \cite{dgd} under the same CPU and GPU settings, respectively. This illustrates that the proposed DHSL is much more efficient than state of the arts.

\section{Conclusion}\label{sec:conclusion}
 In this paper, a deep hybrid similarity learning (DHSL) method for person Re-IDentification (Re-ID) is proposed. The superior performance of DHSL is achieved by reasonably assigning complexities of metric learning and feature learning modules in the CNN model. In the metric learning module, the hybrid similarity function is proposed and learned on the element-wise absolute differences and multiplications of the CNN learning feature pairs simultaneously, which yields a more discriminative similarity metric. In the feature learning module, a light CNN only including three convolution layers is applied. We further examine the effectiveness of each role in the proposed DHSL method, e.g., complementary behavior of element-wise absolute differences and multiplications, training dataset scale and running time analysis. Experiments on three challenging person Re-ID databases, QMUL GRID, VIPeR and CUHK03, show the proposed DHSL method consistently outperforms multiple state-of-the-art person Re-ID methods.
%
%


%



\ifCLASSOPTIONcaptionsoff
  \newpage
\fi



%

\bibliographystyle{IEEEtran}
\small{
\bibliography{paper}
}

\end{document}